\documentclass[runningheads]{llncs}
\usepackage{graphicx}

\usepackage{tikz}
\usepackage{comment}
\usepackage{amsmath,amssymb} 
\usepackage{color}

\usepackage[accsupp]{axessibility}  
\usepackage[width=122mm,left=12mm,paperwidth=146mm,height=193mm,top=12mm,paperheight=217mm]{geometry}

\usepackage{subcaption}
\usepackage{booktabs}
\usepackage{caption}
\usepackage{multirow}
\usepackage{bbm}

\usepackage{xcolor}

\begin{document}
\pagestyle{headings}
\mainmatter
\def\ECCVSubNumber{6093}  

\title{Action Quality Assessment with Temporal Parsing Transformer} 

\titlerunning{ECCV-22 submission ID \ECCVSubNumber} 
\authorrunning{ECCV-22 submission ID \ECCVSubNumber} 
\author{Anonymous ECCV submission}
\institute{Paper ID \ECCVSubNumber}

\titlerunning{Action Quality Assessment with Temporal Parsing Transformer}
%
\author{Yang Bai\inst{1,2*\dagger} \and
Desen Zhou\inst{2*} \and Songyang Zhang\inst{3} \and Jian Wang\inst{2} \and
 Errui Ding\inst{2}, \\ Yu Guan\inst{4} \and Yang Long\inst{1} \and
Jingdong Wang\inst{2\ddagger}}

\authorrunning{Y. Bai et al.}
\institute{
Department of Computer Science, Durham University \and Department of Computer Vision Technology (VIS), Baidu Inc. \and Shanghai AI Laboratory, $^{4}$ University of Warwick \\
\email{\{yang.bai,yang.long\}@durham.ac.uk},\email{\{zhoudesen,wangjingdong\}@baidu.com}}

\maketitle


{\let\thefootnote\relax\footnotetext{
\scriptsize 
* Equal contribution. $\dagger$ Work done when Yang Bai was a research intern at VIS, Baidu. \\ $\ddagger$ Corresponding Author.
}}\par

\begin{abstract}
Action Quality Assessment(AQA) is important for action understanding and resolving the task poses unique challenges due to subtle visual differences. Existing state-of-the-art methods typically rely on the holistic video representations for score regression or ranking, which limits the generalization to capture fine-grained intra-class variation. To overcome the above limitation, we propose a temporal parsing transformer to decompose the holistic feature into temporal part-level representations. Specifically, we utilize a set of learnable queries to represent the atomic temporal patterns for a specific action. Our decoding process converts the frame representations to a fixed number of temporally ordered part representations. To obtain the quality score, we adopt the state-of-the-art contrastive regression based on the part representations. Since existing AQA datasets do not provide temporal part-level labels or partitions, we propose two novel loss functions on the cross attention responses of the decoder: a ranking loss to ensure the learnable queries to satisfy the temporal order in cross attention and a sparsity loss to encourage the part representations to be more discriminative. Extensive experiments show that our proposed method outperforms prior work on three public AQA benchmarks by a considerable margin.

\keywords{action quality assessment, temporal parsing transformer, temporal patterns, contrastive regression}

\end{abstract}

\section{Introduction}

Action quality assessment(AQA), which aims to evaluate how well a specific action is performed, has attracted increasing attention in research community recently~\cite{mltaqa,aqa7}.
In particular, assessing the action quality accurately has great potential in a wide range of applications such as health care \cite{aqahealth} and sports analysis~\cite{bertasius2017baller,pirsiavash2014assessing,parmar2017learning,mltaqa}. 

In contrast to the conventional action recognition tasks~\cite{wang2013action,i3d}, AQA poses unique challenges due to the subtle visual differences. Previous works on AQA either use ranking-based pairwise comparison between test videos\cite{doughty2018s} or estimate the quality score with regression-based methods\cite{parmar2017learning,xu2019learning}. However, these methods typically
represent a video with its \textit{holistic representation}, via the global pooling operation over the output of the backbone network(e.g., I3D\cite{i3d}).  Since the videos to be evaluated usually are from the same coarse action category (e.g., diving) in AQA, it's crucial to capture \textit{fine-grained intra-class variation} to estimate more accurate quality scores. Thus, we propose to decompose the holistic feature into more fine-grained temporal part-level representations for AQA.

\begin{figure*}[t]
\begin{center}
\includegraphics[width=1\linewidth]{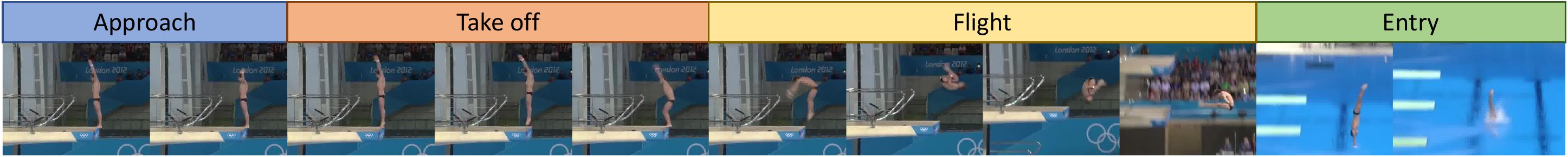}
\end{center}
   \caption{An action consists of multiple temporally ordered key phases.}
\label{fig:intro}
\end{figure*}

To achieve this, a promising strategy is to represent the video by using a set of atomic action patterns. For example, a diving action consists of several key phases, such as \textit{approach}, \textit{take off}, \textit{flight}, etc., as illustrated in Fig.\ref{fig:intro}. The fine-grained patterns enable the model to describe the subtle differences, which is expected to improve the 
assessment of action quality effectively.
Nevertheless, it remains challenging to learn such atomic patterns as the existing AQA datasets do not provide temporal part-level labels or partitions.

In this work, we aim to tackle the aforementioned limitations by developing a regression-based action quality assessment strategy, which enables us to leverage the fine-grained atomic action patterns without any explicit part-level supervision. Our key idea is to model the shared atomic temporal patterns, with a set of learnable queries for a specific action category. Similar to the decoding process of transformer applied in natural language modeling\cite{vaswani2017attention}, we propose a temporal parsing transformer to decode each video into a fixed number of part representations. To obtain quality scores, we adopt the recent state-of-the-art contrastive regression framework\cite{yu2021core}. Our decoding mechanism allows the part representations between test video and exemplar video to be implicitly aligned via a shared learnable query.
Then, we generate a relative pairwise representation per part and fuse different parts together to perform the final relative score regression.

To learn the atomic action patterns without the part-level labels, we propose two novel loss functions on the cross attention responses of the decoder.
Specifically, to ensure the learnable queries satisfy the temporal order in cross attention, we calculate an attention center for each query by weighted summation of the attention responses with their temporal clip orders. Then we adopt a marginal ranking loss on the attention centers to guide the temporal order. Moreover, we propose a sparsity loss for each query's attention distribution to guide the part representations to be more discriminative.

We evaluate our method, named as temporal parsing transformer(TPT), on three public AQA benchmarks: MTL-AQA\cite{mltaqa}, AQA-7\cite{aqa7} and JIGSAWS\cite{jigsaw}. As a result, our method outperforms previous state-of-the-art methods by a considerable margin. The visualization results show that our method is able to extract part-level representations with interpretable semantic meanings. We also provide abundant ablation studies for better understanding.


The main contributions of this paper are three folds:
\begin{itemize}
    \item We propose a novel temporal parsing transformer to extract fine-grained temporal part-level representations with interpretable semantic meanings, which are optimized with the contrastive regression framework.
    \item We propose two novel loss functions on the transformer cross attentions to learn the part representations without the part-level labels.
    \item We achieve the new state-of-the-art on three public AQA benchmarks, namely MTL-AQA, AQA-7 and JIGSAWS.
\end{itemize}

    
    

\section{Related Work}


\subsection{Action quality assessment}
In the past years, the field of action quality assessment (AQA) has been repaid developed with a broad range of applications such as health care\cite{aqahealth}, instructional video analysis\cite{doughty2018s,doughty2019pros}, sports video analysis\cite{aqa7,mltaqa}, and many others\cite{gordon1995automated,jug2003trajectory}. Existing AQA methods can be categorized into two types: regression based methods and ranking based methods.

\paragraph{\textbf{Regression based methods}}
Mainstream AQA methods formulate the AQA task as a regression task based on reliable score labels, such as scores given by expert judges of sports events.
For example,  Pirsiavash et al. \cite{pirsiavash2014assessing} took the first steps towards applying the learning method to the AQA task and trained a linear SVR model to regress the scores of videos based on handcrafted features.
Gordan et al. \cite{gordon1995automated} proposed in their pioneer work the use of skeleton trajectories to solve the problem of quality assessment of gymnastic vaulting movements.
Parmar et al. \cite{parmar2017learning} showed that spatiotemporal features from C3D \cite{c3d} can better encode the temporal representation of videos and significantly improve AQA performance.
They also propose a large-scale AQA dataset and explore all-action models to better evaluate the effectiveness of models proposed by the AQA community.
Xu et al. \cite{xu2019learning} proposed learning multi-scale video features by stacked LSTMs followed \cite{parmar2017learning}.
Pan et al. \cite{pan2019action} proposed using spatial and temporal graphs to model the interactions between joints. Furthermore, they also propose to use I3D \cite{i3d} as a stronger backbone network to extract spatiotemporal features.
Parmar et al. \cite{mltaqa} introduced the idea of multi-task learning to improve the model capacity of AQA, and collected AQA datasets with more annotations to support multi-task learning.
To diminish the subjectiveness of the action score from human judges, Tang et al. \cite{usdl2020} proposed an uncertainty-aware score distribution learning (USDL) framework Recently.
However, the video's final score can only provide weak supervision concerning action quality. Because two videos with different low-quality parts are likely to share similar final scores, which means the score couldn't provide discriminative information.
\paragraph{\textbf{Ranking based methods}}
Another branch formulates AQA task as a ranking problem.
Doughty et al. \cite{doughty2018s} proposed a novel loss function that learns discriminative features when a pair of videos exhibit variance in skill and learns shared features when a pair of videos show comparable skill levels.
Doughty et al. \cite{doughty2019pros} used a novel rank-aware loss function to attend to skill-relevant parts of a given video.
However, they mainly focus on longer, more ambiguous tasks and only predict overall rankings, limiting AQA to applications requiring some quantitative comparisons.
Recently, Yu et al. \cite{yu2021core} proposed the Contrastive Regression (CoRe) framework to learn the relative scores by pair-wise comparison, highlighting the differences between videos and guiding the models to learn the key hints for assessment.

\subsection{Temporal action parsing}
Fine-grained action parsing is also studied in the field of action segmentation or temporal parsing \cite{lea2017temporal,kuehne2014language,alayrac2017joint,lei2018temporal,li2019weakly,yi2021asformer}. For example, Zhang et al.\cite{zhang2021temporal} proposed Temporal Query Network adopted query-response functionality that allows the query to attend to relevant segments. Dian et al.\cite{shao2020intra} proposed a temporal parsing method called TransParser that is capable of mining sub-actions from training data without knowing their labels.
However, different from the above fields,  part-level labels are not available in AQA task. Furthermore, most of the above methods focus more on frame-level feature enhancement, whereas our proposed method extracts part representations with interpretable semantic meanings.


\section{Method}
In this section, we introduce our temporal parsing transformer with the contrastive regression framework in detail.


\begin{figure*}[t]
\begin{center}
\includegraphics[width=1\linewidth]{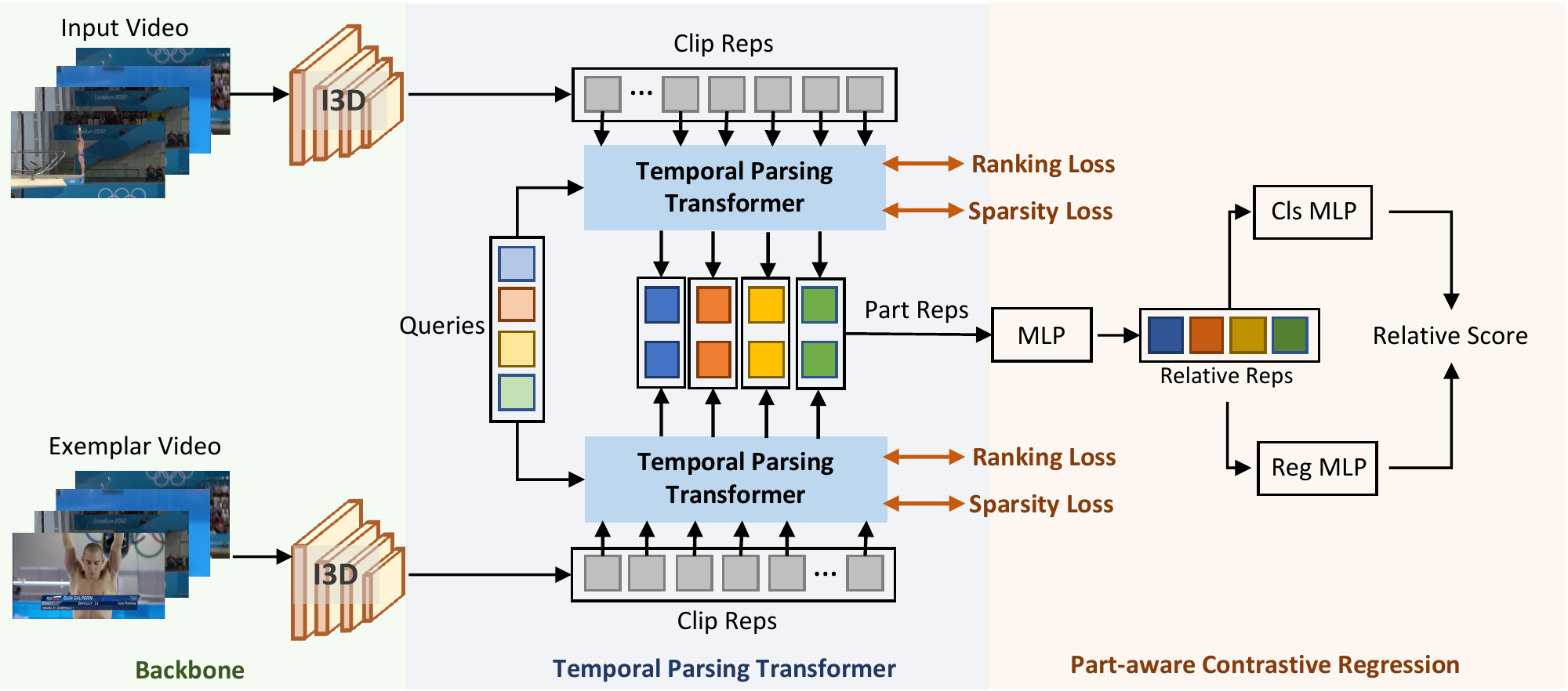}
\end{center}
   \caption{\small Overview of our framework. Our temporal parsing transformer converts the clip-level representations into temporal part-level representations. Then the part-aware contrastive regressor first computes part-wise relative representations and then fuses them to estimate the relative score. We adopt the group-aware regression strategy, following\cite{yu2021core}.
   During training, we adopt the ranking loss and sparsity loss on the decoder cross attention maps to guide the part representation learning.}
\label{fig:overview}
\end{figure*}

\subsection{Overview}
The input of our network is an action video. We adopt the Inflated 3D ConvNets(I3D)\cite{i3d} as our backbone, which first applies a sliding window to split the video into $T$ overlapping clips, where each clip contains $M$ consecutive frames. Then, each clip goes through the I3D network, resulting in time series clip level representations $\boldsymbol{V} = \{\boldsymbol{v}_t \in \mathbb{R}^{D}\}^T_{t=1}$, where $D$ is feature dimension and $T$ is the total number of clips. In our work, we do not explore spatial patterns, hence each clip representation $\boldsymbol{v}_t$ is obtained by average pooling across spatial dimensions. The goal of AQA is to estimate a quality score $\mathbf{s}$ based on the resulting clips representation $\boldsymbol{V}$. In contrastive regression framework, instead of designing a network to directly estimate raw score $\mathbf{s}$, it estimates a relative score between the test video and an exemplar video $\boldsymbol{V}_0$ with known quality score $\mathbf{s}_0$, which is usually sampled from training set. Then, contrastive regression aims to design a network $\mathcal{F}$ that estimates the relative score $\Delta\boldsymbol{s}$:
\begin{equation}
     \Delta \mathbf{s} = \mathcal{F}(\boldsymbol{V}, \boldsymbol{V}_0),
\end{equation}
then final score can be obtained by
\begin{equation}
    \mathbf{s} = \mathbf{s}_0 + \Delta \mathbf{s}.
\end{equation}
In our framework, we first adopt a temporal parsing transformer $\mathcal{G}$ to convert the clip level representations $\boldsymbol{V}$ into temporal part level representations, denoted by $\boldsymbol{P} = \{\boldsymbol{p}_k\in \mathbb{R}^d\}_{k=1}^K$, where $d$ is the part feature dimension and $K$ is the number of queries, i.e. temporal atomic patterns. Then for test video and exemplar video, we can have two set of aligned part representations $\boldsymbol{P}$ and $\boldsymbol{P}_0=\{\boldsymbol{p}_k^0\in \mathbb{R}^d\}_{k=1}^K$. Our new formulation can be expressed as:
\begin{equation}
    \Delta \mathbf{s} = \mathcal{R}(\boldsymbol{P}, \boldsymbol{P}_0).
\end{equation}
where $\mathcal{R}$ is the relative score regressor, and
\begin{equation}
    \boldsymbol{P} = \mathcal{G}(\boldsymbol{V}), \boldsymbol{P}_0 = \mathcal{G}(\boldsymbol{V}_0).
\end{equation}
An overview of our framework is illustrated in Fig.\ref{fig:overview}.
Below we describe the detailed structure of temporal parsing transformer $\mathcal{G}$ and part-aware contrastive regressor $\mathcal{R}$.
\subsection{Temporal parsing transformer}
Our temporal parsing transformer takes the clip representations as memory and exploits a set of learnable queries to decode part representations. Different from prevalent DETR architecture\cite{detr}, our transformer only consists of a decoder module. We found that the encoder module does not provide improvements in our framework; it even hurts the performance. We guess it might because that clip-level self-attention smooths the temporal representations, and our learning strategy cannot decode part presentations in this way without part labels. 

We perform slight modifications to the standard DETR decoder. That is, the cross attention block in our decoder has a learnable parameter, temperature, to control the amplification of the inner product. Formally, in the $i$-th decoder layer, the decoder part feature $\{p_k^{(i)}\in\mathbb{R}^d\}$ and learnable atomic patterns(i.e. query set) $\{q_k\in\mathbb{R}^d\}$ are first summed as a query and then perform cross attention on the embedded clip representation $\{\mathbf{v}_t\in \mathbb{R}^{d}\}$:
\begin{equation}
    \alpha_{k,t} = \frac{\exp^{(\boldsymbol{p}_k^{(i)}+q_k)^T\cdot \boldsymbol{v}_t/\tau}}{\sum\limits_{j=1}^{T} \exp^{(\boldsymbol{p}_k^{(i)}+q_k)^T\cdot\boldsymbol{v}_j/\tau}},
    \label{eq:atten}
\end{equation}
where $\alpha_{k,t}$ indicates the attention value for query $k$ to clip $t$, $\tau\in \mathbb{R}$ indicates the learnable temperature to enhance the inner product to make the attentions more discriminative. Unlike DETR\cite{detr}, in our decoder, we do not utilize position embedding of clip id to the memory $\{\boldsymbol{v}_t\}$. We expect our query to represent atomic patterns, instead of spatial anchors, as in the detection task\cite{anchordetr,conditionaldetr}. We found that adding position encoding significantly drops the performance and makes our learning strategy fail, which will be shown in the experiment section. 

In our experiments, we only utilize one-head attention in our cross attention blocks. The attention values are normalized across different clips, since our goal is to aggregate clip representations into our part representation. Then the updated part representation $\boldsymbol{p}_{k}^{(i)'}$ has the following form:
\begin{equation}
    \boldsymbol{p}_{k}^{(i)'} = \sum\limits_{j=1}^{T} \alpha_{k,j}\boldsymbol{v}_j+ \boldsymbol{p}_{k}^{(i)}.
\end{equation}
We then perform standard FFN and multi-head self-attention on decoder part representations. Similar to DETR\cite{detr}, our decoder also has a multi-layer structure.

\subsection{Part-aware contrastive regression}
Our temporal parsing transformer converts the clip representations $\{\boldsymbol{v}_t\}$ into part representations $\{\boldsymbol{p}_k\}$. Given a test video and exemplar video, we can obtain two part representation sets $\{\boldsymbol{p}_k\}$ and $\{\boldsymbol{p}_k^0\}$. One possible way to estimate the relative quality score is to fuse each video's part representations and estimate the relative score. However, since our temporal parsing transformer allows the extracted part representations to be semantically aligned with the query set, we can compute the relative pairwise representation per part and then fuse them together. Formally, we utilize a multi-layer perceptron(MLP) $f_r$ to generate the relative pairwise representation $\boldsymbol{r}_k\in\mathbb{R}^d$ for $k$-th part:
\begin{equation}
    \boldsymbol{r}_k = f_r(Concat([\boldsymbol{p}_k;\boldsymbol{p}_k^0])).
\end{equation}
The MLP $f_r$ is shared across different parts. To balance the score distributions across the whole score range, we adopt the group-aware regression strategy to perform relative score estimation\cite{yu2021core}. Specifically, it first calculates $B$ relative score intervals based on all possible pairs in training set, where each interval has equal number of pair-samples. Then it generates a one-hot classification label $\{l_n\}$, where $l_n$ indicates whether the ground truth score $\Delta \mathbf{s}$ lies in $n$-th interval, and a regression target $\gamma_n = \frac{\Delta\mathbf{s}-x_{left}^n}{x_{right}^n-x_{left}^n}$, where $x_{left}^n, x_{right}^n$ denote the left and right boundary of $n$-th interval. Readers can refer to \cite{yu2021core} for more details.

We adopt average pooling\footnote[1]{
We note that it might be better to weight each part. However, part weighting does not provide improvements during our practice. We guess that it may be during the self-attention process in the decoder, the relations between parts have already been taken into account.
} on the relative part representations $\{\boldsymbol{r}_k\}$ and then utilize two two-layer MLPs to estimate the classification label $\{l_n\}$ and regression target $\{\gamma_n\}$. Different from \cite{yu2021core}, we do not utilize tree structure. Since we have obtained fine-grained part-level representations and hence the regression becomes simpler, we found that two-layer MLP works fine.
 
\subsection{Optimization}
Since we do not have any part-level labels at hand, it's crucial to design proper loss functions to guide the part representation learning. We have assumed that each coarse action has a set of temporally ordered atomic patterns, which are encoded in our transformer queries. To ensure that our query extracts different part representations, we constrain the attention responses in cross attention blocks for different queries. Specifically, in each cross attention process, we have calculated the normalized attention responses $\{\alpha_{k,t}\}$ by Eq.\ref{eq:atten}, then we compute an attention center $\Bar{\alpha}_k$ for $k$-th query: 
\begin{equation}
    \Bar{\alpha}_k = \sum\limits_{t=1}^T t\cdot \alpha_{k,t},
    \label{eq:center}
\end{equation}
where $T$ is the number of clips and $\sum\limits_{t=1}^T \alpha_{k,t} = 1$. Then we adopt two loss functions on the attention centers: ranking loss and sparsity loss.
\paragraph{\textbf{Ranking loss}}
To encourage that each query attends to different temporal regions, we adopt a ranking loss on the attention centers. We wish our part representations have a consistent temporal order across different videos. To this end, we define an order on the query index and apply ranking losses to the corresponding attention centers. We exploit the margin ranking loss, which results in the following form:
\begin{equation}
    L_{rank} =\sum_{k=1}^{K-1} \max(0, \Bar{\alpha}_k-\Bar{\alpha}_{k+1} + m) + \max (0, 1- \Bar{\alpha}_1 + m) + \max(0, \Bar{\alpha}_K - T + m),
    \label{eq:ranking}
\end{equation}
where $m$ is the hyper-parameter margin controlling the penalty, the first term guides the attention centers of part $k$ and $k+1$ to keep order: $\Bar{\alpha}_{k} < \Bar{\alpha}_{k+1}$. From Eq. \ref{eq:center}, we have the range of attention centers: $1 \leq \Bar{\alpha}_{k} \leq T$. To constrain the first and last part where $k=1$ and $k=K$, we assume there is two virtual centers at boundaries: $\Bar{\alpha}_0=1$ and $\Bar{\alpha}_{K+1}=T$. The last two terms in Eq. \ref{eq:ranking} constrain the first and last attention centers not collapsed to boundaries.

\paragraph{\textbf{Sparsity loss}}
To encourage the part representations to be more discriminative, we further propose a sparsity loss on the attention responses. Specifically, for each query, we encourage the attention responses to focus on those clips around the center $\mu_{k}$, resulting in the following form:

\begin{equation}
    L_{sparsity} = \sum\limits_{k=1}^{K} \sum\limits_{t=1}^{T} |t - \Bar{\alpha}_{k}| \cdot \alpha_{k,t}
\end{equation}
During training, our ranking loss and sparsity loss are applied to the cross attention block in each decoder layer.

\paragraph{\textbf{Overall training loss}}
In addition to the above auxiliary losses for cross attention, our contrastive regressor $\mathcal{R}$ generates two predictions for the group classification label $\{l_n\}$ and regression target $\{\gamma_n\}$, we follow \cite{yu2021core} to utilize the BCE loss on each group and square error on the ground truth regression interval:
\begin{equation}
    L_{cls} = -\sum\limits_{n=1}^{N} l_n \log (\tilde{l}_n) + (1-l_n) \log(1-\tilde{l}_n)
\end{equation}
\begin{equation}
    L_{reg} = \sum\limits_{n=1}^{N} \mathbbm{1}(l_n=1) (\gamma_n - \tilde{\gamma}_n)^2
\end{equation}
where $L_{reg}$ only supervises on the ground truth interval, $\tilde{l}_n$ and $\tilde{\gamma}_{n}$ are predicted classification probability and regression value. The overall training loss is given by:
\begin{equation}
    L_{all} = \lambda_{cls} L_{cls} + \lambda_{reg} L_{reg} + \lambda_{rank} \sum\limits_{i=1}^{L}L_{rank}^{i} + \lambda_{sparsity}\sum\limits_{i=1}^{L} L_{sparsity}^{i},
\end{equation}
where $i$ indicates layer id and $L$ is the number of decoder layers, $\lambda_{cls}$, $\lambda_{reg}$, $\lambda_{rank}$, $\lambda_{sparsity}$ are hyper-parameter loss weights.

\section{Experiment}
\subsection{Experimental Settup}
\paragraph{\textbf{Datasets}} 
We perform experiments on three public benchmarks: MTL-AQA\cite{mltaqa},  AQA-7\cite{aqa7}, and JIGSAWS\cite{jigsaw}. 
See Supplement for more details on datasets.

\paragraph{\textbf{Evaluation Metrics}} 
Following prior work\cite{yu2021core}, we utilize two metrics in our experiments, the Spearman’s rank correlation and relative L2 distance($R\mbox{-}\ell_2$). \textbf{Spearman's rank correlation} was adopted as our main evaluation metric to measure the difference between true and predicted scores.
The Spearman’s rank correlation is deﬁned as follows:
\begin{equation}
    \rho = \frac{\sum_i (p_i - \bar{p})(q_i - \bar{q})}{\sqrt{\sum_i (p_i - \bar{p})^2\sum_i (q_i - \bar{q})^2}}
\end{equation}
It focuses on the ranking of test samples. In contrast, \textbf{relative L2 distance} measures the numerical precision of each sample compared with ground truth. Formally, it's defined as:
\begin{equation}
    R\mbox{-}\ell_2 = \frac{1}{N}\sum_{n = 1}^{N}(\frac{|s_n - \hat{s}_n|}{s_{max}-s_{min}})^2
\end{equation}

\subsubsection{Implementation Details}
We adopt the I3D backbone pretrained on Kinetics\cite{i3d} as our local spatial-temporal feature extractor. 
The Adam optimizer is applied with a learning rate $1 \times 10^{-4}$  for the backbone 
and transformer module. The learning rate for the regression head is set to $1 \times 10^{-3}$. 
The feature dimension is set to 512 for the transformer block. We select 10 exemplars for each test sample during the inference stage to align with previous work\cite{yu2021core} for fair comparisons. 
As for the data-preprocessing on AQA-7 and MTL-AQA datasets, we sample 103 frames following previous works for all videos. Since our proposed method requires more fine-grained temporal information, unlike previous work that segmented the sample frames into 10 clips, we segment the frames into 20 overlapping clips each containing 8 continuous frames. As for the JIGSAWS dataset, we uniformly sample 160 frames following \cite{usdl2020} and divide them into 20 non-overlapping clips as input of the I3D backbone. We select exemplars from the same difficulty degree on MTL-dataset during the training stage. For AQA-7 and JIGSAWS datasets, all exemplars come from the same coarse classes.

\begin{table}[t]
\caption{\small Performance comparison on MTL-AQA dataset. `w/o DD' means that training and test processes do not utilize difficulty degree labels, `w/ DD' means experiments utilizing difficulty degree labels.}
    \centering
     \resizebox{0.45\textwidth}{!}{
        \begin{tabular}{l c c }
        \hline
        Method (w/o DD) &
        Sp. Corr. & 
        $R\mbox{-}\ell_2(\times 100)$   \\ 
        \hline
        Pose+DCT\cite{pirsiavash2014assessing} & 0.2682 & - \\
        C3D-SVR\cite{parmar2017learning} & 0.7716 & - \\
        C3D-LSTM\cite{parmar2017learning} & 0.8489 & - \\
        MSCADC-STL\cite{mltaqa} & 0.8472 & - \\
        C3D-AVG-STL\cite{mltaqa} & 0.8960 & - \\
        MSCADC-MTL\cite{mltaqa} & 0.8612 & - \\
        C3D-AVG-MTL\cite{mltaqa} & 0.9044 & - \\
        USDL\cite{usdl2020} &0.9066 & 0.654 \\
        CoRe\cite{yu2021core} & 0.9341 & 0.365  \\
        TSA-Net\cite{wang2021tsa} & 0.9422 & - \\
        Ours & \textbf{0.9451} & \textbf{0.3222} \\
        \hline
        \hline
        Method (w/ DD) & Sp. Corr & $R\mbox{-}\ell_2$($\times$ 100) \\
        \hline
        USDL\cite{usdl2020} &0.9231 & 0.468 \\
        MUSDL\cite{usdl2020} &0.9273 & 0.451 \\
        CoRe\cite{yu2021core} & 0.9512 & 0.260  \\
        Ours & \textbf{0.9607} & \textbf{0.2378} \\
        \hline
        \end{tabular}
        }
        \label{tb:MTL}
\end{table}

\begin{table}[t]
\caption{\small Performance comparison on AQA-7 dataset.}
    \centering
     \resizebox{0.95\textwidth}{!}{
        \begin{tabular}{l  c c  c c  c c  c }
        \hline
        Sp. Corr &
        Diving &
        Gym Vault &
        BigSki. &
        BigSnow. &
        Sync. 3m &
        Sync. 10m    &
        Avg. Corr. \\
        \hline
        Pose+DCT\cite{pirsiavash2014assessing} &0.5300 &0.1000 &- &- &- &- &-\\
        ST-GCN\cite{stgcn} &0.3286 &0.5770 &0.1681 &0.1234 &0.6600 &0.6483 &0.4433\\
        C3D-LSTM\cite{parmar2017learning} &0.6047 &0.5636 &0.4593 &0.5029 &0.7912 &0.6927 &0.6165\\
        C3D-SVR\cite{parmar2017learning} &0.7902 &0.6824 &0.5209 &0.4006 &0.5937 &0.9120 &0.6937\\
        JRG\cite{pan2019action} &0.7630 &0.7358 &0.6006 &0.5405 &0.9013 &0.9254 &0.7849\\
        USDL\cite{usdl2020} &0.8099 &0.7570 &0.6538 &0.7109 &0.9166 &0.8878 &0.8102\\
        CoRe\cite{yu2021core} & 0.8824 & 0.7746 & 0.7115 & 0.6624 & 0.9442 & 0.9078 & 0.8401 \\
        TSA-Net\cite{wang2021tsa}                 &0.8379 &0.8004 &0.6657 &0.6962 &\textbf{0.9493} &0.9334 &0.8476\\
        Ours & \textbf{0.8969} & \textbf{0.8043} &\textbf{0.7336} &\textbf{0.6965} &0.9456 & \textbf{0.9545} &\textbf{0.8715} \\
        \hline
        \hline
        $R\mbox{-}\ell_2(\times 100)$ & Diving & Gym Vault & BigSki.& BigSnow. & Sync. 3m & Sync. 10m & Avg. $R\mbox{-}\ell_2$\\
        \hline
        C3D-SVR\cite{parmar2017learning} & 1.53 & 3.12 & 6.79 & 7.03 & 17.84 & 4.83 & 6.86 \\
        USDL\cite{usdl2020} & 0.79 & 2.09 & 4.82 & 4.94 & 0.65 & 2.14 & 2.57\\
        CoRe\cite{yu2021core} & 0.64 & 1.78 & 3.67 & 3.87 & 0.41 & 2.35 & 2.12\\
        Ours &\textbf{0.53} &\textbf{1.69} &\textbf{2.89} &\textbf{3.30} &\textbf{0.33} & \textbf{1.33} &\textbf{1.68} \\
        \hline
        \end{tabular}
        }
        \label{tb:aqa-7}
\end{table}

\begin{table}[t]
\caption{\small Performance comparison on JIGSAW dataset.}
    \centering
     \resizebox{0.4\textwidth}{!}{
        \begin{tabular}{l  c c c c }
        \hline
        
        Sp. Corr. & S & NP & KT & Avg. \\
        \hline
        ST-GCN\cite{stgcn} & 0.31 & 0.39 & 0.58 & 0.43 \\
        TSN\cite{wang2016temporal} & 0.34 & 0.23 & 0.72 & 0.46 \\
        JRG\cite{pan2019action} & 0.36 & 0.54 & 0.75 & 0.57 \\
        USDL\cite{usdl2020} & 0.64 & 0.63 & 0.61 & 0.63 \\
        MUSDL\cite{usdl2020} & 0.71 & 0.69 & 0.71 & 0.70 \\
        CoRe\cite{yu2021core} & 0.84 & 0.86 & 0.86 & 0.85 \\
        Ours & \textbf{0.88} & \textbf{0.88} & \textbf{0.91} & \textbf{0.89} \\
        \hline
        \hline
        $R\mbox{-}\ell_2 $ & S & NP & KT & Avg. \\
        \hline
        CoRe\cite{yu2021core} & 5.055 & 5.688 & 2.927 & 4.556 \\
        Ours & \textbf{2.722} & \textbf{5.259} & \textbf{3.022} & \textbf{3.668} \\
        \hline
        \end{tabular}
        }
        \label{tb:jigsaw}
\end{table}

\subsection{Comparison to state-of-the-art}
We compare our results with state-of-the-art methods on three benchmarks in Tab.\ref{tb:MTL}, Tab.\ref{tb:aqa-7} and Tab.\ref{tb:jigsaw}. Our method outperforms priors works on all three benchmarks under all settings. 

On \textbf{MTL-AQA dataset}, we evaluated our experiments with two different settings, following prior work\cite{yu2021core}. Specifically, the MTL-AQA dataset contains the label of difficult degree, and each video's quality score is calculated by the multiplication of the raw score with its difficulty. 
In the experiment setting `w/o DD', the training and test processes do not utilize difficulty degree labels. In setting `w/ DD', we exploit the difficulty label by comparing the test video to the exemplar videos with the same difficulty, and we estimate the raw score, which is multiplied by the difficulty to get the final quality. Our method outperforms existing works under both settings. As shown in Tab. \ref{tb:MTL}, under `w/ DD', our method achieves a Sp. Corr. of 0.9607, and $R\mbox{-}\ell_2$ of 0.2378, outperforms the tree-based CoRe\cite{yu2021core}. Note that our model simply utilizes two shallow MLPs to perform contrastive regression instead of the tree structure as in \cite{yu2021core}. Our transformer extracts fine-grained part representations, hence the regression becomes easier. Under `w/o DD', out method achieves 0.9451(Sp. Corr) and 0.3222($R\mbox{-}\ell_2$), outperforms the CoRe and recent TSA-Net\cite{wang2021tsa}. It's worth noting that TSA-Net utilizes an external VOT tracker\cite{vot-toolkit} to extract human locations and then enhance backbone features, which is orthogonal to the main issue of temporal parsing addressed in our work. Consequently, we expect that our method can be further improved by incorporating the attention module as in \cite{wang2021tsa}.


On \textbf{AQA-7  dataset}, our method achieves state-of-the-art on 5 categories and comparable performance on the rest category, shown in Tab. \ref{tb:aqa-7}. In particular, on average, our method outperforms CoRe by 3.14 Corr.($\times 100$) and TSA-Net by 2.39 Corr.($\times 100$), and obtains a very small $R\mbox{-}\ell_2$ of 1.68($\times 100$), demonstrating the effectiveness of our temporal parsing transformer.

On the smallest \textbf{JIGSAW dataset}, we perform 4-fold cross validation for each category, following prior work\cite{yu2021core,usdl2020}. Our method achieves an average of 0.89 Corr. and 3.668 $R\mbox{-}\ell_2$, achieves new state-of-the-art.

\begin{table}[t]
\caption{\small Ablation study of different components on MTL-AQA dataset.}
    \centering
     \resizebox{0.55\textwidth}{!}{
        \begin{tabular}{c c c c | c c }
        \hline
        Method &
        TPT & $L_{rank}$ & $L_{sparsity}$ & Sp. Corr. & $R\mbox{-}\ell_2$ \\
        \hline
        Baseline & $\times$ & $\times$ & $\times$ & 0.9498 & 0.2893 \\
         & \checkmark & $\times$ & $\times$ & 0.9522 & 0.2742 \\
         & \checkmark & \checkmark & $\times$ & 0.9583 & 0.2444 \\
        Ours & \checkmark & \checkmark & \checkmark & \textbf{0.9607} & \textbf{0.2378} \\
        \hline
        \end{tabular}
        }
        \label{tb:components}
\end{table}

\subsection{Ablation Study}
In this subsection, we perform ablation studies to evaluate the effectiveness of our proposed model components and designs. All of our ablation studies are performed on MTL-AQA dataset under `w/ DD' setting. We build a baseline network that directly pool the clip features without transformer, and utilize the resulting holistic representation to perform contrastive regression. 

\paragraph{\textbf{Different model components}}
In this work, we propose a novel temporal parsing transformer(TPT), and exploit the ranking loss($L_{rank}$) and sparsity loss($L_{sparsity}$) on cross attention responses to guide the part representation learning. We first perform experiments to show the effectiveness of each design, the results are shown in Tab.\ref{tb:components}. We can observe that with only TPT, the performance only improves marginally from 0.9498 Corr. to 0.9522 Corr.. With the ranking loss, the performance is significantly improved, demonstrating the importance of temporally ordered supervision strategy. The sparsity loss further improves the performance, showing that the discrimination of parts is also important.

\begin{table}[t]
    \caption{\small More ablation studies on MTL-AQA dataset.}
    
    \begin{subtable}{.45\linewidth}
      \centering
        \caption{\small Different part generation strategies.}
        \label{tb:ablation_parsing}
        \resizebox{0.9\textwidth}{!}{
        \begin{tabular}{l c c c c c }
        \hline
        Method & Sp. Corr. & $R\mbox{-}\ell_2$ \\
        \hline
        Baseline & 0.9498 & 0.2893 \\
        Adaptive pooling & 0.9509 & 0.2757 \\
        Temporal conv & 0.9526 & 0.2758 \\
        TPT(ours) & \textbf{0.9607} & \textbf{0.2378} \\
        \hline
        \end{tabular}}
    \end{subtable}
    \begin{subtable}{.48\linewidth}
      \centering
        \caption{\small Effect of order guided supervision.}
        \label{tb:ablation_diversity}
        \resizebox{0.9\textwidth}{!}{
        \begin{tabular}{c | c | c }
        \hline
        Method & Sp. Corr. & $R\mbox{-}\ell_2$ \\
        \hline
        Baseline & 0.9498 & 0.2893 \\
        Diversity loss & 0.9538 & 0.2655 \\
        Ranking loss(ours) & \textbf{0.9607} & \textbf{0.2378} \\
        \hline
        \end{tabular}
        }
    \end{subtable}
    \begin{subtable}{1.01\linewidth}
      \centering
        \caption{\small Effect of positional encoding.}
        \label{tb:ablation_pos}
        \resizebox{0.7\textwidth}{!}{
        \begin{tabular}{c c c | c c }
        \hline
        Pos. Encode & Memory(clip) & Query(part) & Sp. Corr. & $R\mbox{-}\ell_2 $ \\
        \hline
        & \checkmark & \checkmark  & 0.9526 & 0.2741 \\
       & \checkmark & $\times$ &  0.9532 & 0.2651 \\
        Proposed & $\times$ & $\times$ & \textbf{0.9607} & \textbf{0.2378} \\
        \hline
        \end{tabular}
        }
    \end{subtable}%
    
    \begin{subtable}{1\linewidth}
      \centering
        \caption{\small Different relative representation generation.}
        \label{tb:ablation_fusion_stage}
        \begin{tabular}{c |c |c }
            \hline
            Method & Sp. Corr. & $R\mbox{-}\ell_2$ \\
            \hline
            Baseline & 0.9498 & 0.2893 \\
            Part-enhanced holistic & 0.9578 & 0.2391 \\
            Part-wise relative + AvgPool(ours) & \textbf{0.9607} & \textbf{0.2378} \\
            \hline
        \end{tabular}
    \end{subtable}%
\end{table}


\paragraph{\textbf{Different relative representation generation}} Since we have obtained part representations  from TPT for each video, we may have two options to generate relative representation for contrastive regression. For the first option, we can first fuse the part representations with a pooling operation for each video, then each video takes the part-enhanced holistic representation to estimate the relative score. For the second option, which is our proposed strategy, we first compute a part-wise relative representation and then apply the AvgPool operation over the parts. We compare the results of above options in Tab.\ref{tb:ablation_fusion_stage}. We can see that the part-wise strategy outperforms part-enhanced strategy. It's worth noting that the part-enhanced approach also outperforms our baseline network, which implies that each part indeed encodes fine-grained temporal patterns.
\paragraph{\textbf{Different part generation strategies}} Our method utilizes the temporal parsing transformer to extract part representations. In this ablation study, we compare our method with the other two baseline part generation strategy, shown in Tab. \ref{tb:ablation_parsing}. The first strategy utilizes the adaptive pooling operation cross temporal frames to down-sample the origin $T$ clip representation into $K$ part representations. The second strategy replaces the above adaptive pooling with a temporal convolution with stride $\lfloor T/K\rfloor$, resulting in a representation with $K$ size. We found that both strategies introduce minor improvements as they can not capture fine-grained temporal patterns.

\paragraph{\textbf{Effect of position encoding}} Different from conventional transformer\cite{detr,vaswani2017attention}, our transformer decoding process does not rely on the temporal position encoding. We compare the results of different position encoding strategies on the memory(clip) and query(part) in Tab.\ref{tb:ablation_pos}. To embed the position encoding on queries, we add the cosine series embedding of $\lfloor T/K \rfloor \times i$ to $i$-th learnable query, making the queries have positional guidance uniformly distributed across temporal clips. We keep the ranking loss and sparsity for fair comparisons. From Tab.\ref{tb:ablation_pos}, we can observe that adding position encoding hurts the learning of temporal patterns.
\paragraph{\textbf{Effect of order guided training strategy}} Our ranking loss on the attention centers consistently encourages the temporal order of atomic patterns. To verify the importance of such order guided supervision, we replace the ranking loss to a diversity loss following the Associative Embedding\cite{newell2017associative} to push attention centers: $L_{div} = \sum\limits_{i=1}^K\sum\limits_{j=i+1}^K \exp^{-\frac{1}{2\sigma^2}(\Bar{\alpha}_i-\Bar{\alpha}_j)^2}$. Compared with $L_{rank}$, $L_{div}$ does not encourage the order of queries, but keeps diversity of part representations. As shown in Tab. \ref{tb:ablation_diversity}, the performance significantly drops from 0.9607 Corr. to 0.9538 Corr., demonstrating the effectiveness of our order guided training strategy.

\begin{figure*}[t]
\begin{center}
\includegraphics[width=1\linewidth]{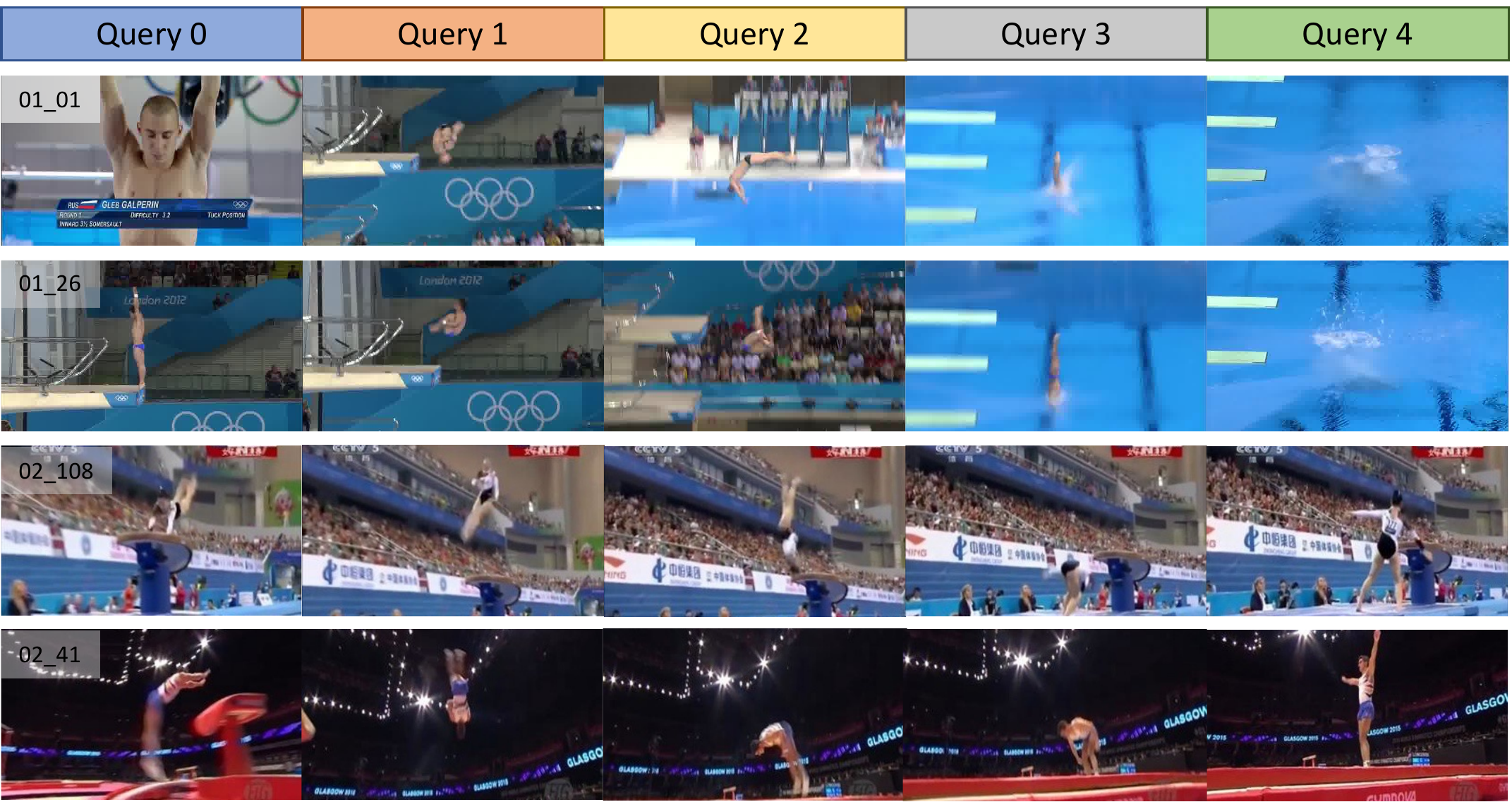}
\end{center}
   \caption{\small 
 Visualization of the frames with the highest attention responses in decoder cross attention maps on MTL-AQA and AQA-7 datasets. Each row represents a test video from different representative categories (diving from MTL-AQA, gymnastic vault from AQA-7), whose ID is shown in the left first frame. Different columns correspond to temporally ordered queries. The above results show that our transformer is able to capture semantic temporal patterns with learned queries.
   }
\label{fig:vis_video}
\end{figure*}

\begin{figure*}[!ht]
\begin{center}
\includegraphics[width=1\linewidth]{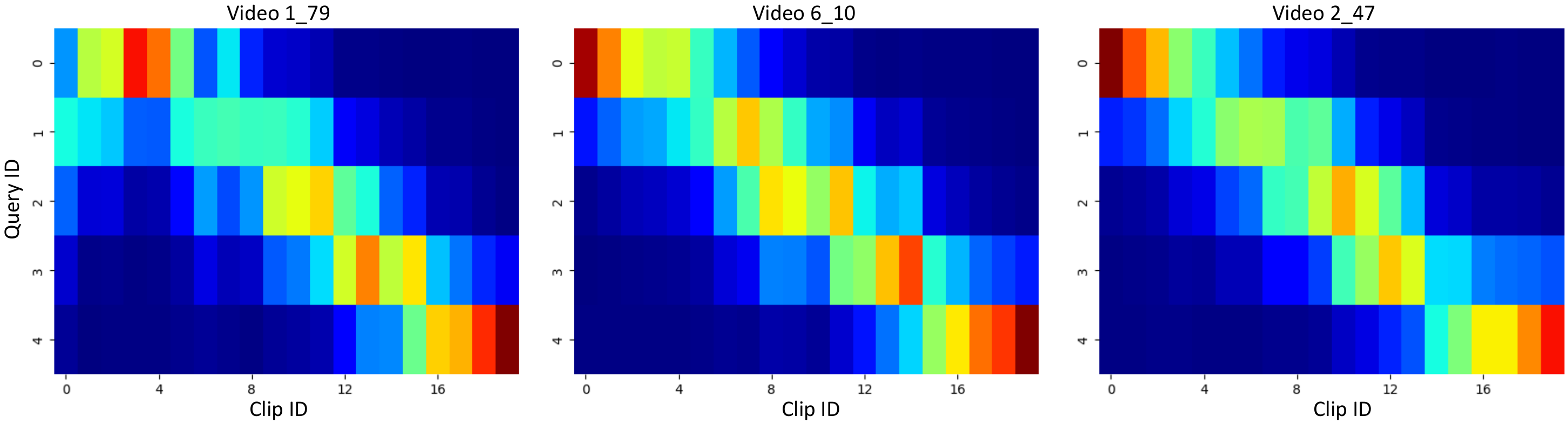}
\end{center}
   \caption{\small Visualization of cross attention maps on three video samples from MTL-AQA dataset, where video IDs are shown on the top. In each subfigure, each row indicates one query, and each column indicates one clip. We can observe that the bright grids(with high attention responses) have a consistent temporal order due to ranking loss, and the attention maps are sparse due to our sparsity loss.}
\label{fig:vis_attention}
\end{figure*}

\subsection{Visualization results}
We provide some visualization results in Fig.\ref{fig:vis_video} and Fig.\ref{fig:vis_attention}. Samples are from MTL-AQA dataset trained under `w/ DD' setting and AQA-7 dataset. In Fig.\ref{fig:vis_video}, we visualize the clip frames with the highest attention responses in cross attention maps of the last decoder layer. Since each clip consists of multiple frames, we select the middle frame of a clip as representative. We can observe that our transformer can capture semantic temporal patterns with learned queries. 
In Fig.\ref{fig:vis_attention}, we visualize the cross attention maps. We can observe that the attention responses have a consistent temporal order due to our designed ranking loss, and they are also sparse due to our sparsity loss. 

\section{Conclusion}
In this paper, we propose a novel temporal parsing transformer for action quality assessment. We utilize a set of learnable queries to represent the atomic temporal patterns, and exploit the transformer decoder to
convert clip-level representations to part-level representations. To perform quality score regression, we exploit the contrastive regression framework that first computes the relative pairwise representation per part and then fuses them to estimate the relative score. To learn the atomic patterns without part-level labels, we propose two novel loss functions on cross attention responses to guide the queries to attend to temporally ordered clips. As a result, our method is able to outperform existing state-of-the-art methods by a considerable margin on three public benchmarks. The visualization results show that the learned part representations are semantic meaningful.



\newpage

\bibliographystyle{splncs04}
\bibliography{aqa}
\end{document}